# Automatic Detection of Epileptiform Discharges in the EEG

André Rosado

Agostinho Rosa

*Evolutionary Systems and Biomedical Engineering Laboratory (LASEEB)*
*ISR-IST-UTL, Lisbon*
arosado@laseeb.org
acrosa@laseeb.org

*Abstract*— The diagnosis of epilepsy generally includes a visual inspection of EEG recorded data by the Neurologist, with the purpose of checking the occurrence of transient waveforms called interictal epileptiform discharges. These waveforms have short duration (less than 100 ms), so the inspection process is usually time-consuming, particularly for ambulatory long-term EEG records. Therefore, an automatic detection system of epileptiform discharges can be a valuable tool for a Neurology service.
The proposed approach is the development of a multi-stage detection algorithm, which processes the complete EEG signals and applies decision criteria to selected waveforms. It employs EEG analysis techniques such as Wavelet Transform and Mimetic Analysis, complemented with a classification based on Fuzzy Logic.
In order to evaluate the algorithm's performance, data were collected from several epileptic patients, with epileptiform activity marked by a Neurologist. The average values obtained for both Sensitivity and Specificity were respectively higher than 80% and 70%.

## I.  Introduction

Electroencephalography is a non-invasive technique, which makes use of various electrodes in the scalp in order to capture electrophysiological activity originated in the brain. The product of these measures is the Electroencephalogram (EEG), which consists in the set of recorded signals in each channel. The EEG has been a valuable tool in the diagnosis and evaluation of neurological disorders, in particular for epilepsy. Epilepsy is characterized by sudden recurrent and transient disturbances of mental function and/or movements of the body that result from excessive discharging of groups of brain cells [1]. Patients who are suspected of having the disease are always subject to an EEG recording.

The presence of epileptiform activity, which is distinct from the background EEG, confirms the diagnosis of epilepsy, although the waveform pattern can be confused with the one resulting from other disorders. During seizures, the scalp EEG of patients is characterized by high-amplitude and synchronized periodic EEG waveforms. These periods are known as ictal activity. In the interictal period (i.e. between seizures), it is typically observed a specific type of transient waveforms, called epileptiform discharges. Interictal activity includes mainly spikes and sharp waves. It is generally accepted that spikes and sharp waves have a high correlation with seizure occurrence [1]. Therefore, the detection of spikes in the EEG plays a key role in the diagnosis of epilepsy, supporting the Neurologist in his evaluation tasks.

A spike wave is usually distinguishable from the background EEG activity by visual inspection. It has a pointed peak and duration of 20 to 70 ms, while a sharp wave as a similar morphology, but longer duration (up to 200 ms). Few detection algorithms make a distinction between spikes and sharp waves [2], so the general term "spikes" encompasses both epileptiform discharges. Epileptiform discharges waveforms have a high level of variability between patients and even on the same record, which constitutes an obstacle to automatic detection. Other difficulty is caused by the presence of artifacts in the EEG signals. This terminology refers to the noise-like phenomena in the signals, either originated by technical defects in the acquisition, or from other physiological sources, such as eye blinks, muscle activity or heart beating interference in the EEG signals. The removal of artifacts is one of the main performance requirements in the detection of epileptiform discharges.

Besides the routine EEG (1-hour duration), it is usual the need for long-term recordings (up to 24 hours), whether made in an EEG laboratory, or through the means of an ambulatory device (AEEG). Especially in the latter cases, the visual inspection of EEG recording made by the Neurologist is very time-consuming, thus raising the need for an automated detection system for epileptiform discharges. There are already some commercial software applications for this purpose, and several algorithms have been developed in this field. Some systems employ a technique called Mimetic Analysis, which reproduces the visual process of the physician, by making the measurements of parameters of EEG waveforms, such as sharpness, slope, duration and amplitude and comparing them with thresholds, which are representative of typical true spikes [3,4]. Other relevant methodologies used include: Artificial Neural Networks [5,7], Expert System [6,8,9], Inverse Filtering [10], and Wavelet Transform [11,12]. The proposed methodology consisted in a multi-stage detection algorithm, following an approach similar with Pattern Recognition systems. The main purpose was to identify and collect candidate waveforms (considered as events) in the complete signals, which were then classified as either epileptiform or non-epileptiform. Initially, the Wavelet Transform was employed for processing and filtering of relevant waveforms in an EEG signal. Then, a feature extraction phase was performed based on Mimetic Analysis, by decomposing the selected waveforms in relevant parameters. Finally, the classification of the waveforms was based on Fuzzy Logic, which was complemented by defining also a set of heuristic rules for post-classification. The results were compared against the detections made by a Neurologist. In this line, an event can be included on one of the four possible categories:

- True Positive (TP): Event marked by the system as epileptiform, and confirmed by the expert;
- False Positive (FP): Event marked by the system as epileptiform, but not confirmed by the expert;
- True Negative (TN): Event marked by the system as non-epileptiform, and confirmed by the expert;
- False Negative (FN): Event marked by the system as non-epileptiform, but not confirmed by the expert.

These measures can be related between each other for the calculation of Sensitivity and Specificity rates, and the overall performance of the algorithm was based on the calculation of these rates for a set of EEG files.

$$\text{Sensitivity} = TP/(TP+FN) \quad (1)$$
$$\text{Specificity} = TN/(TN+FP) \quad (2)$$

## II. EEG PRE-PROCESSING USING THE WAVELET TRANSFORM

### A. General Properties of the Wavelet transform

The Wavelet Transform (WT) is a tool used in the context of signal processing, performing a time-frequency analysis on a time-series, which reveals its non-stationarities. Where the Fourier Transform (FT) provides an analysis solely in terms of frequency components of the signal, the WT can also provide time information of a given section, guaranteeing good resolution in both domains. For this reason, the WT constitutes a valuable option for EEG analysis, being adequate to the non-stationary properties of the signals.

The WT reveals the details of a signal in a scale-frequency context, allowing the data to be analyzed at different scales or resolutions. These scales can be compared with windows through which we can look at the data: large windows correspond to less detail, while small windows correspond to fine resolutions with a lot of detail. Furthermore, they relate in an inverse manner to the frequency: the smaller the scale, the higher the frequencies to be highlighted. The Continuous Wavelet Transform (CWT) is defined by:

$$CWT(\tau, a) = \int_{-\infty}^{+\infty} x(t) \psi_{\tau,a}^{*}(t) dt \quad (3)$$

In this relation, $\psi(t)$ is the wavelet function, given by:

$$\psi_{\tau,a}(t) = \frac{1}{\sqrt{|a|}} \psi\left(\frac{t-\tau}{a}\right) \quad (4)$$

In practice, the transform is a convolution of the original signal $x(t)$, with a mother wavelet $\psi(t)$. The mother wavelet is a basis function defined on a finite domain, making it appropriate to approximate non-stationary data like the EEG waveforms. The mother wavelet can be dilated and translated to fit the signal, according to the parameters $a$ and $\tau$, respectively. The result of a wavelet transformation is a set of coefficients that represent the similarity of the signal with the dilated and translated versions of the mother wavelet. A proper choice of the mother wavelet is crucial for the analysis and a good option is to take in account the shape of the events to be detected. So in this context, the main idea of CWT is to evaluate the correspondence of the mother wavelet with a given signal in multiple scales.

### B. Multiresolution analysis for EEG spike detection

The CWT allows performing a multiresolution analysis in the EEG signals, with the objective to detect spike waveform candidates. Multiresolution analysis refers to the examination of signals at different resolution scales. The detection process uses this technique to assess the differences between epileptiform and the background activity in an EEG segment, along the various scales of analysis (Fig. 1 and Fig. 2).

The application of the CWT in some scales produces coefficient vectors with maximum values correspondent to the spike waves peaks in the EEG segment. This behavior was taken in account, and with the addition of a peak detection mechanism in the CWT coefficients vector, it was possible to localize the candidate waveforms in each EEG segment.

In this approach, certain options had to be made in respect to CWT parameters, namely the choices of mother wavelet $\psi(t)$ and scale, and for that purpose it was performed preliminary tests to calculate Sensitivity for different values of these parameters. The same records were used for both parameters' tests, each one with a significant number of epileptiform events marked by an expert, and also different acquisition conditions in terms of sampling rate. For these tests, the Sensitivity rate was calculated with a simple decision criteria based on the establishment of a threshold value for the coefficients of each segment analyzed. This is in line the objective of filtering and selecting the most relevant events of each signal for the subsequent stages.

It must be added that for the CWT processing, each EEG signal is segmented in 10-second windows in a "sliding window" mode. For the choice of the mother wavelet it was defined a test to assess the results for five different wavelet functions (Table I), which resulted in the selection of

Daubechies-2 function (Fig. 3) for the optimization of the algorithm.

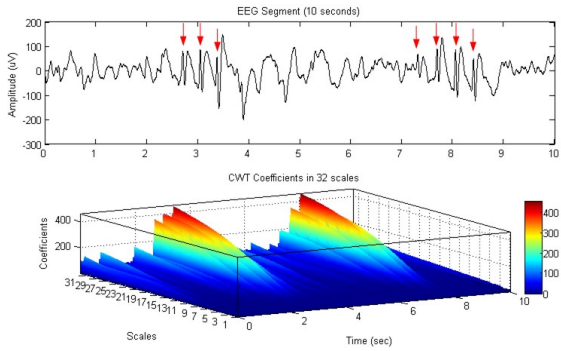

Fig. 1: Time-scale analysis performed by the CWT application on a 10-seconds segment with epileptiform activity (marked on the EEG segment)

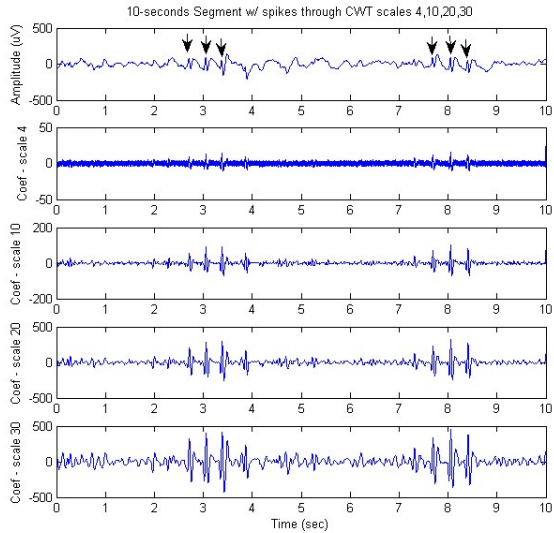

Fig. 2: Multiresolution analysis performed by CWT (scales:4,10,20 and 30) on an EEG segment with 10 seconds and containing epileptiform discharges (marked on the segment)

Besides the wavelet function, it was also assessed the performance for Sensitivity rate in presence of different analysis scales (Table II). The scale of CWT also influences considerably the results in terms of detection rate. Each resolution level will result in a different behavior for a given segment (Fig.2), which is easily explained given the meaning of the variation of the parameter. The conclusions taken from this test were not so direct in this case, as each scale provides significant differences in performance. Because of the sampling rate variation in each record, the temporal resolutions of each waveform are different, and consequently the detection rate will be substantially different for some EEG records. This observation resulted in the option of setting different scale values according to the temporal resolution (or sampling rate) of a signal.

Taking in account the options made for CWT parameters, it was established the threshold for filtering candidate epileptiform events in the EEG segments. The threshold value is proportional to the standard deviation of the analyzed EEG segment, thus being linked with the statistical properties of data. Depending on the level of precision desired for this initial stage, it can be more or less selective in the filtering of events. As a general remark, it was shown that the CWT-based detection is able to effectively reduce the original data in a meaningful set for the classification phase, with Sensitivity values near 100% for a non-conservative threshold.

TABLE I. MOTHER WAVELET INFLUENCE ON SENSITIVITY RATE

| Mother Wavelet | Sensitivity (%) | | |
|---|---|---|---|
| | Record 1 (695 spikes) | Record 2 (601 spikes) | Record 3 (224 spikes) |
| Daubechies-2 | 95,53 | 89,68 | 95,98 |
| Daubechies-4 | 85,90 | 83,69 | 75,89 |
| Daubechies-5 | 77,99 | 82,52 | 76,33 |
| Coiflet-4 | 85,90 | 85,52 | 84,38 |
| Symmlet-8 | 86,04 | 86,18 | 86,60 |

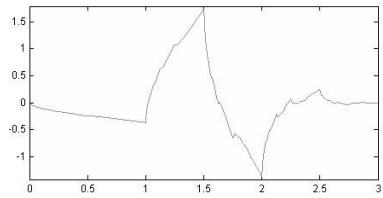

Fig. 3: Daubechies-2 wavelet function

III. FEATURE EXTRACTION

Mimetic Analysis refers to the technique of reproducing the behavior and decision process of an expert in an automated way. The basis of this approach in the EEG signal processing field is the interval-amplitude analysis. This procedure consists in decomposing waveforms in half-waves, defining them in duration and in amplitude [1]. For the latter, the values are obtained by the difference between maximum and minimum values of the wave, and in terms of duration, it can be obtained from the sum of the two half-waves, after a division from the peak. This analysis is more effective with a larger temporal resolution in the signal, which means more samples for each waveform.

The decomposition provides the basis for the mimetic analysis, by reducing the waveform to a set of descriptive parameters, which will then be subject to a decision based on criteria similar to the Neurologist´s. The classification phase will then be dependent on the set of features chosen by this approach, which also contributes to the definition of rules.

In the feature extraction procedure, the parameters are calculated for each waveform selected by the CWT, and are grouped as attributes of that particular event. Thus, a feature vector descriptive of the waveform will be used as input to the classifier. This vector includes significant information regarding the event and the criteria for the parameter's choice was built from the knowledge obtained from experts and other related work [3]. Given the morphology of a spike wave (Fig.

4), the parameters presented on Table III were extracted as features of an epileptiform event.

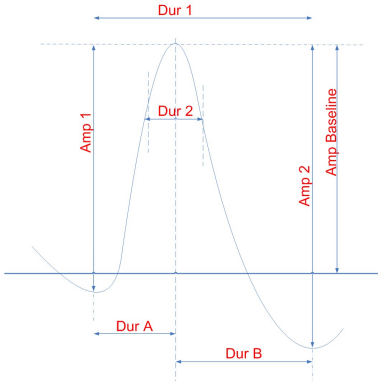

Fig. 4: Representation of a spike wave and the correspondent decomposition and feature extraction

TABLE II.    FEATURES OF EPILEPTIFORM DISCHARGES

| Feature | Description |
| --- | --- |
| AMP1 | Amplitude measured from the start of 1st half-wave to the peak, in µV |
| AMP2 | Amplitude measured from the peak to the end of 2nd half-wave, in µV |
| AMP baseline | Amplitude measured from the peak to the reference level, in µV |
| DURA | Duration of the 1st half-wave, in ms |
| DURB | Duration of the 2nd half-wave, in ms |
| DUR1 | Total Duration of the waveform, in ms |
| DUR2 | Duration measured between the points with maximum slope in each half-have, in ms |
| SLOPE1 | Slope of the 1st half-wave |
| SLOPE2 | Slope of the 2nd half-wave |

The amplitudes of both half-waves and the baseline allow a differentiation from the EEG background activity, because interictal activity has higher values than some of its most characteristics rhythms (e.g. alpha, beta, theta). However, in the case of EEG waveforms having values similar to interictal discharges, it is necessary to take in account other parameters like duration or slope for a correct classification. This comparative analysis is analogous to the visual inspection made by the EEG experts, which is the basis for the chosen methodology and is reflected in the choice of the features for the classification phase.

## IV. CLASSIFICATION OF EPILEPTIFORM EVENTS

### A. General Concepts of Fuzzy Logic

The use of Fuzzy Logic brings added value in the specification of the classifier, because of the possibility to introduce expert knowledge on the waveforms features and the output classes. Fuzzy Logic has the property to be adaptable to more subjective and imprecise data, which is the case for EEG signals. Moreover, it makes a correspondence to a class as a matter of degree, not on absolute terms.

At a general level, Fuzzy Logic consists in the mapping of the input space to the output by the means of inference rules, which are formulated in terms of *IF-THEN* propositions. The mechanism to create the inference rules is based on fuzzy sets and membership functions. Fuzzy sets differentiate from the usual ones by not having a clearly defined boundary (e.g.: a given amplitude can have a correspondence to both sets "small" and "medium"), while a membership function is a curve that defines how each point in the input space is mapped to a membership value (or degree of membership) between 0 and 1. If $X$ is the universe of discourse and its elements are denoted by $x$, then a fuzzy set $A$ in $X$ is defined as a set of ordered pairs:

$$A = \{(x, \mu_A(x)) \mid x \in X\} \quad (5)$$

In this relation $\mu_A(x)$ is the membership function of $x$ in $A$. The inference rules are written in the notation "*IF x is A THEN y is B*", where A and B are fuzzy sets. The final stage of the classifier is the defuzzification, which produces the output of the classification, converting the sum of all rules (the aggregate output fuzzy set) in a single number.

### B. Specification of the Fuzzy Logic Classifier

The outcome of the feature extraction phase will be the input to the classification, and each *n*-input to the classifier will consist in a vector with the *n*-value of each parameter. The classifier was specified with the objective of being independent of the EEG record and analyzed signal, by having broad margins for the fuzzy sets, assuming the variability of each parameter value. As the main purpose is to correctly identify the epileptiform discharges, the decision process was modeled according to the known characteristics of these waveforms. The "Amplitude 1" and "Amplitude 2" have similar membership functions, as well as "Duration A" and "Duration B". For these pairs of features, the inference rules establish relation criteria that are typical in spike waves (e.g.: generally Amp 2 is greater than Amp 1). The range of admissible values was defined to be larger than the limit of what is considered to be normal value for EEG waveforms (< 400 µV). Some other options were taken during the training phase of the classifier, by realizing that there are values which although are part of marked events by the expert, are outside the usual intervals. One important issue is the temporal resolution of the signal as influence on certain parameters. More specifically, the "Duration" margins have to consider the case of a sample representing 10 ms of the signal (i.e. 100 samples/sec), which for typical spike durations (20-70 ms), introduces an associated error in the measures after decomposition.

The output value was defined to be in the interval [0,1], and also to have a *large* value when the event is considered to be epileptiform (Fig. 5). The *small* and *medium* intervals correspond to the situation of, respectively, and non-epileptiform event and a possible one. This last class was introduced to guarantee a certain confidence interval to the decision, and allow the reviewer to reject or confirm the event. On the other hand, the events that are >= 0.8 will be marked by the automatic system as epileptiform.

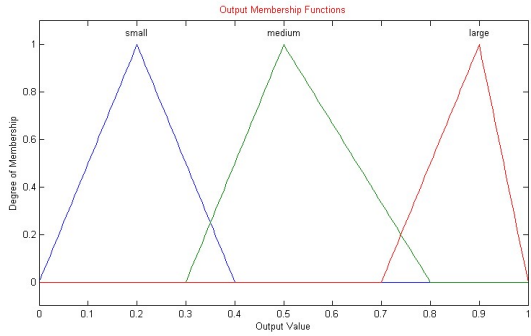

Fig. 5: Membership Functions for output classes

The formulation of inference rules is the part of specification where it is more clearly made use of empirical knowledge of the epileptiform discharges characteristics, in order to discriminate them from the remaining waveforms. This information is represented in a symbolic way [13], similar to description of an expert during the evaluation, and allows the correspondence between input data and the defined output classes. The set of rules was derived based on spike waves parameters properties, although most of them were adjusted during the training phase, in order for the classification to achieve a greater consistency, as it was observed that the waveforms values vary significantly between records.

### C. Post-Classification

With the purpose to refine and improve global performance of the automatic detection, it was also introduced a set of heuristic rules, aimed at eliminating false positives (FP). A large value of FP is a major drawback for every detection system, and in the case of EEG analysis and epileptiform activity, it is generally caused by artifacts. The objective is then for post-classification to complement fuzzy classification: discarding an event in the case of an erroneous detection, or otherwise assuming the prior value obtained by the classification.

The strategy for elimination of FP is based on traditional characteristics of artifacts, and also by employing temporal context analysis. For instance, an event separated by a short interval (< 100 ms) from a previous positive one is discarded, considering the expected temporal distance between spikes and/or sharp events in a signal. It is also taken in account the features of other FP which are not artifacts, such as sleep events (K complexes, spindles or vertex waves), and EEG rhythms like alpha. It must also be noted that the heuristic rules for post-classification (Table III) have a different context than the inference rules described in Fuzzy Logic context. The justification is that for eliminating FP, the rules are mainly built in terms of rejection criteria, which is based on known threshold values for certain parameters.

TABLE III. POST-CLASSIFICATION RULES

| Applicable Feature | Description | Target FP to reject |
|---|---|---|
| AMP1, AMP2 | <50μV | Alpha rhythm |
| DUR1 | <20 ms | EMG and alpha rhythm |
| DUR1 | >350 ms | K Complex |
| DURA, DURB | >150 ms | EOG |

## V. RESULTS

A set of records from epileptic patients was gathered for validation of the system, with a subset coming from sleep records with 8 or less channels, and other from routine EEG with 16 channels (see Table IV). The calculation of the results followed the condition of only considering an event as TP if it was in an interval below 50 ms of a marked event, and if inside that timeframe there was more than one detection, all of them were grouped together. After the calculation of TP and the other events' categories, the Sensitivity and Specificity were calculated for each record. Given that the classifier output parameter is in the range [0,1], different values for both rates can be determined by varying a threshold inside that interval. In that case, a more adequate analysis can be made, using the ROC curve ("*Receiver Operating Characteristic*"). The ROC curve captures the various performance rates available for each decision threshold in one graph, allowing a comparative characterization for each output value. It also provides, for a given record, the value that optimizes the detector both in terms of Sensitivity and Specificity.

TABLE IV. SET OF RECORDS FOR VALIDATION (16 CHANNELS)

| Record No. | Duration (min) | No. marked events |
|---|---|---|
| 3 | ~80 | 601 |
| 5 | ~85 | 387 |
| 6 | ~75 | 320 |

### A. Results for Fuzzy Logic classification

Table V presents the results for Sensitivity and Specificity in three records, while Fig. 6 displays the correspondent ROC overlapped curves. Fuzzy Logic classification achieves a good performance, being consistent over 80% in terms of Sensitivity, and also with Specificity above 80% in one record, with another ~70%, for optimal values. The reference level observed in Fig. 6 is the common level of performance that has to be surpassed in terms of ROC analysis in order to obtain valid detection results, which it is accomplished in all cases considered for these evaluation tests. An important outcome from the results was that the optimal values in each ROC curve are coincident with the most selective threshold (0.8), which is also the highest confidence level that the system delivers to the user.

TABLE V. PERFORMANCE RESULTS FOR 3 RECORDS FOR FUZZY CLASSIFIER

| Output Threshold | Record 3 | | Record 5 | | Record 6 | |
|---|---|---|---|---|---|---|
| | Sens(%) | Spe(%) | Sens(%) | Spe(%) | Sens(%) | Spe(%) |
| >0,2 | 93,3 | 17,1 | 99,0 | 0,4 | 99,1 | 0 |
| >0,3 | 93,0 | 45,6 | 88,1 | 67,1 | 98,8 | 8,8 |
| >0,4 | 93,0 | 47,0 | 87,9 | 69,1 | 98,8 | 9,6 |
| >0,5 | 92,7 | 49,6 | 87,3 | 73,3 | 95,0 | 25,2 |
| >0,6 | 92,5 | 53,7 | 85,8 | 75,5 | 90,3 | 33,2 |
| >0,7 | 92,3 | 55,2 | 85,0 | 79,0 | 83,1 | 50,5 |
| >0,8 | 91,5 | 56,6 | 84,8 | 81,4 | 77,8 | 68,7 |

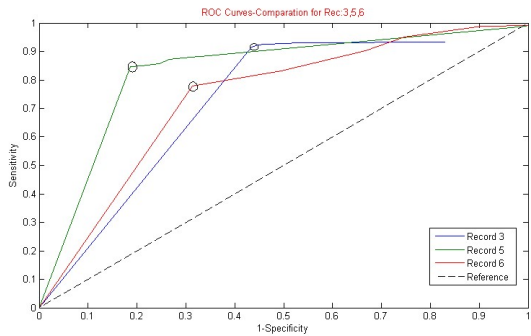

Fig. 6: ROC Curves for comparing performance in each record, with each optimization value pointed out

### B. Results with post-classification

In the post-classification tests, it is intended to check the improvement in performance by including rules for post-classification in the detection system. Particularly, the values of Specificity for records 3 and 6 on Table VI indicate the presence of too many FP, which are intended to be minimized with this complementary stage. For the tests with post-classification, the number of threshold values for the classifier output was reduced, given the interest in mainly observing the behavior of the detector for more selective criteria. The results in Table VII show an improvement in Specificity in both records. Specifically, for record 3, there is an increase in Specificity to values higher than 75%, with residual loss of Sensitivity. The comparison of ROC curves in this record (see Fig. 7), with and without post-classification, allows concluding on the performance enhancement with post-classification.

TABLE VII. PERFORMANCE RESULTS FOR 2 RECORDS W/POST-CLASSIFICATION

| Output Threshold | Record 3 | | Record 6 | |
|---|---|---|---|---|
| | Sens(%) | Spe(%) | Sens(%) | Spe(%) |
| >0,5 | 89,2 | 76,1 | 86,3 | 57,0 |
| >0,6 | 89,2 | 76,1 | 85,6 | 58,6 |
| >0,7 | 89,0 | 76,7 | 80,0 | 65,0 |
| >0,8 | 88,4 | 77,0 | 77,5 | 69,9 |

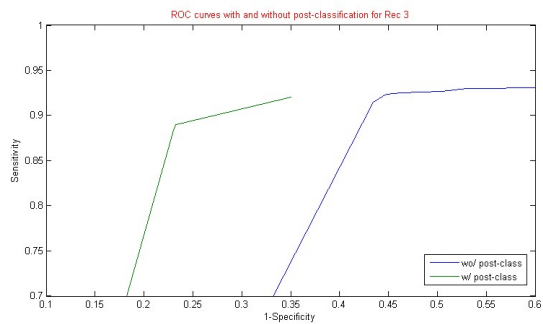

Fig. 7: Comparative ROC curves with (green line) and without (blue line) post-classification for Record 3. The ROC curves were zoomed to enhance visualization of the compared values of Sensitivity and Specificity.

## VI. CONCLUSIONS

The general approach of the proposed methodology was to follow the analysis steps taken by an expert in his decision process, which guided the option of using Mimetic Analysis concepts and classification with Fuzzy Logic. These techniques allowed the incorporation of expert knowledge in decomposition of the spike waveforms in representative parameters, and on the formulation of decision criteria (i.e. inference rules). These options constituted a suitable option for the complex process of EEG classification, where there is a high degree of variability in the waveforms in each record.

CWT signal processing is a very useful tool in the context of EEG analysis, given the non-stationary properties of the signals. In particular, the detection of epileptiform activity can be enhanced by the choice of an appropriate wavelet function for the signal decomposition. This of course can be extrapolated for other EEG waveforms, which highlights the potential of using CWT for different types of analysis (e.g. an artifact detector for specific EEG sleep patterns). The main contribution of CWT in this context is the possibility to filter a first set of relevant events that includes almost 100% of the ones marked by the Neurologist, thus significantly reducing the original data. Each channel can contain several hours of data, so compress it in a group of events in the magnitude of 100-200 ms enhances the efficiency of the classification made by the system.

Fuzzy Logic proved to be a valuable alternative to the more traditional classification methods (e.g. Neural networks), with the advantage that the specification can be very intuitive and easily adjustable. The results obtained for this classifier are clearly in line with state of the art [2], in terms of Sensitivity and Specificity rates. The post-classification stage is an add-on to the global specification, because the set of rules has the only goal to eliminate FP, not contributing to the detection of epileptiform discharges. Nevertheless, his role in that particular task proved to be very important for a system to be applicable in EEG clinical practice.

The overall results obtained for the automatic detection system are considered to be of practical use for EEG laboratories, because of the conjunction of high Sensitivity (>80%) with a very good level of Specificity, so rejecting the majority of FP.

The analysis of ROC curves show that the curves are always above the reference level that establishes the usefulness of a detection system. Also, the optimal value in each curve corresponds to a high classification threshold, which is representative of the associated confidence level.